\documentclass[fleqn,10pt]{wlscirep}
\usepackage[utf8]{inputenc}
\usepackage[T1]{fontenc}
\usepackage{graphicx}
\usepackage{multirow}
\usepackage{amsmath,amssymb,amsfonts}
\usepackage{amsthm}
\usepackage{mathrsfs}
\usepackage[title]{appendix}
\usepackage{xcolor}
\usepackage{textcomp}
\usepackage{manyfoot}
\usepackage{booktabs}
\usepackage{algorithm}
\usepackage{algorithmicx}
\usepackage{algpseudocode}
\usepackage{listings}
\usepackage{tabularx, rotating} 
\usepackage{makecell}

\title{Visual Information Extraction from Documents via Classification-Guided Large Vision-Language Models}

\author[1,*,+]{Huafu Li}
\author[1]{Guo Chen}
\author[1]{Jia Xia}
\author[1]{Lei Wang}
\author[1]{Wei Du}
\author[1]{Yun Yao}
\author[1]{Weijun Peng}
\author[2,+]{Liming Li}

\affil[1]{China Mobile Information Technology Co., Ltd., Shenzhen, 518000, China}
\affil[2]{School of Information Science and Engineering, NingboTech University, Ningbo, 315100, China}

\affil[*]{lihuafu@chinamobile.com}

\affil[+]{These authors contributed equally to this work.}


\begin{abstract}
Visual information extraction (VIE) from visually rich documents remains challenging due to high layout variability and real-world impairments. Existing methods typically rely on sequential OCR pipelines or end-to-end models requiring extensive labeled data and layout-specific training, limiting their scalability.
We propose a classification-guided large vision-language model (LVLM) framework for multi-type VIE that achieves high accuracy with minimal supervision. The approach decouples document-type classification from content extraction and employs in-context learning (ICL)-based dynamic prompt engineering to inject task-specific knowledge, enabling robust zero-shot inference across diverse layouts.  
From a theoretical perspective, the proposed method can be viewed as a form of conditional computation that reduces task uncertainty and improves information efficiency during prompt-based inference.
Evaluated on a real-world bidding dataset with 16 certificate types, our zero-shot method (based on Qwen2.5-VL-7B) outperforms a strong supervised baseline by 18.35 percentage points in F1-score (86.43\% vs. 68.08\%) and 0.23 in normalized edit distance (0.90 vs. 0.67). Optional domain-specific fine-tuning further improves performance to 93.65\% F1 and 0.93 NED, demonstrating superior robustness against seals, watermarks, and low contrast.
The framework offers an efficient, scalable solution for complex document understanding in office automation. Code is available at https://github.com/FairmeHIT/Multi-VIE, and fine-tuned models at https://huggingface.co/fairme/Qwen2.5-VL-7B-SFT. 
\end{abstract}
\begin{document}

\flushbottom
\maketitle

\thispagestyle{empty}

\section*{Introduction}
Visually rich document images -- such as business licenses, financial statements, contracts, tickets, and invoices -- serve as crucial information carriers in diverse application scenarios, containing abundant data essential for business and decision-making processes.
Visual information extraction (VIE) aims to identify and extract predefined semantic entities from these images, forming a foundational component of document understanding and office automation systems \cite{PICK_9412927, review_lin2023visual, ren2025tablegpt}.

Most existing VIE methods adopt a two-stage pipeline: optical character recognition (OCR) for text detection and recognition, followed by natural language processing or unified information extraction (UIE) for entity extraction \cite{wei2024general_ocr2_0, lu2022unified_UIE}. While numerous solutions target these stages individually, integrating them efficiently remains challenging. Visually rich documents incorporate multimodal elements (e.g., layout, fonts, charts, and colors) and real-world noise (e.g., watermarks, seals, blur, and distortions), complicating extraction \cite{VIE_kuang2023visual, review_lin2023visual}.

Deep learning has driven significant progress, with multimodal models integrating semantic, visual, and layout features achieving strong performance in practical applications \cite{VIE_kuang2023visual, VIE_wang2021towards, VIE_zhang2022dual}. However, these approaches typically require large labeled datasets and task-specific training, with evaluations often limited to single-layout documents and few entity types. In real-world settings, extracting information from multiple heterogeneous document types simultaneously (termed multi-VIE) poses a key challenge.

The emergence of large vision-language models (LVLMs), such as GPT-4V \cite{GPT_4V_shi2023exploring} and the Qwen2-VL series \cite{wang2024qwen2}, has sparked interest in their application to VIE. Yet, general-purpose LVLMs exhibit limitations in complex document tasks, including non-Latin script recognition, table understanding, fine-grained perception, and robustness to perturbations, often generating hallucinations \cite{GPT_4V_shi2023exploring, SeeingisBelieving, DianJin-ocr}.

Current frameworks lack robust multi-type real-world evaluation and minimally supervised solutions for diverse layouts. We address these gaps by proposing a classification-guided LVLM framework that achieves high-accuracy structured extraction with minimal task-specific training.

The main contributions of this work are summarized as follows:
\begin{itemize}
    \item We propose a classification-guided conditioning paradigm for LVLM-based multi-VIE, which formulates document understanding as a modular decomposition of task uncertainty into document-type prediction and conditional generation, providing a principled alternative to monolithic prompting.
    \item We introduce a dynamic prompt construction strategy based on predicted class, which serves as an implicit conditioning mechanism that improves task relevance and reduces context redundancy without modifying LVLM parameters.
    \item We provide a theoretical analysis that interprets prompt construction as conditional computation, showing how relevance-aware prompting improves information efficiency, mitigates attention dilution, and enhances in-context learning alignment.
    \item Extensive experiments on real-world and public benchmarks demonstrate the effectiveness, robustness, and generalizability of the proposed framework under both zero-shot and fine-tuned settings.
\end{itemize}

\section*{Related Work}
Traditional VIE methods for scene-specific documents often rely on rule-based strategies, leveraging fixed layouts for dictionary lookup, regular expressions, or pattern matching \cite{review_lin2023visual, schuster2013intellix}. While effective in constrained settings, these approaches require substantial manual adaptation across diverse scenarios, limiting scalability and robustness.

Early deep learning methods advanced VIE by incorporating multimodal features. Grid-based representations, such as Chargrid \cite{katti2018chargrid}, BERTgrid \cite{denk2019bertgrid}, and ViBERTgrid \cite{lin2021vibertgrid}, fuse text and layout in 2D grids but struggle with complex semantic relationships. Graph neural networks (GNNs), including GraphIE \cite{qian2018graphie}, PICK \cite{PICK_9412927}, MatchVIE \cite{tang2021matchvie}, and GraphDoc \cite{zhang2022multimodal}, model inter-component dependencies effectively yet suffer from over-smoothing and training instabilities.

Transformer-based architectures have emerged as dominant, dynamically fusing semantic, visual, and layout features. Representative works include LayoutLM \cite{huang2022layoutlmv3}, StrucTexT \cite{li2021structext}, UDoc \cite{gu2021unidoc}, LayoutXLM \cite{xu2021layoutxlm}, and LiLT \cite{wang2022lilt}, offering strong generalization at the cost of high computation. End-to-end models like EATEN \cite{guo2019eaten}, TRIE \cite{zhang2020trie}, and Donut \cite{kim2022ocr} integrate detection, recognition, and extraction to minimize error propagation, though they typically demand extensive labeled data. Few-shot methods \cite{cheng2020one, wang2022towards} and unified information extraction frameworks \cite{lu2022unified_UIE} address data scarcity, enabling efficient adaptation across tasks and domains.

The rise of large vision-language models (LVLMs) has opened new possibilities for VIE through techniques like chain-of-thought \cite{chen2023chain} and in-context learning (ICL) \cite{cai2023context}. These approaches enhance multimodal reasoning via carefully designed prompts and demonstrations. However, general-purpose LVLMs exhibit limitations in complex document tasks, including OCR for non-Latin scripts, table understanding, and robustness to real-world perturbations \cite{GPT_4V_shi2023exploring}. 
Recent studies have explored advanced prompt learning strategies to improve robustness and generalization. For instance, prompt-based meta-learning methods have been proposed to mitigate label noise and enhance adaptability under distribution shifts \cite{prompt_meta_learning_2026}. Other works investigate automated prompt optimization or task-specific prompt tuning to better align model outputs with desired objectives.
Despite these advances, most existing approaches focus on optimizing prompts in a static or task-agnostic manner, without explicitly considering input-dependent task variation. In complex document understanding scenarios, where multiple document types coexist, a single prompt or globally optimized prompt often fails to capture type-specific extraction requirements.

Recent specialized document LVLMs, such as DocLLM \cite{docllm2024}, TextMonkey \cite{liu2024textmonkey}, and the Monkey series \cite{monkey2024,li2025monkeyocr}, incorporate layout-aware designs and OCR-free processing, achieving strong performance on structured documents. Despite these advances, they often rely on task-specific fine-tuning or prompts and struggle with zero-shot adaptation to highly variable multi-type documents without dynamic, classification-guided knowledge injection -- the gap addressed by our work.

\section*{Methodology}
\label{sec:methodology}
To tackle the multi-VIE challenge, we propose a classification-guided LVLM framework that achieves high accuracy and generalization with minimal task-specific training. This serves as our primary contribution. For comparison, we also present an enhanced OCR\&UIE pipeline that requires more supervised training but provides a strong trainable baseline. We describe the main LVLM-based framework first, followed by the alternative OCR\&UIE approach.

\subsection*{Classification-Guided LVLM Framework for Multi-VIE} 
\label{lvlm_based_framework} 
\begin{figure*}[h] 
  \centering
  \includegraphics[width=1\linewidth]{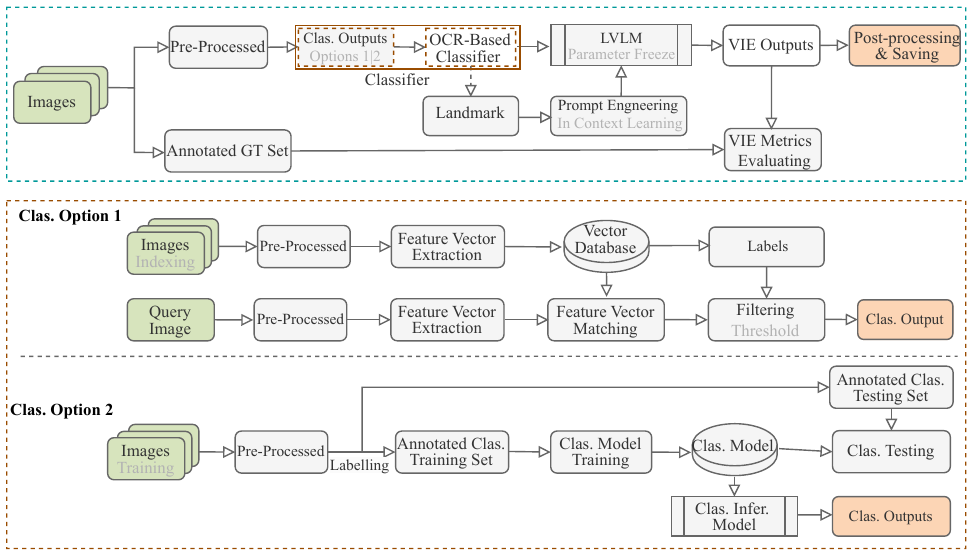}  
  \caption{Overview of the classification-guided LVLM framework for multi-VIE. The design leverages a parameter-frozen LVLM to support minimally supervised, scalable deployment.
  }
  \label{fig:LVLM_based} 
\end{figure*}

Our core framework leverages the pre-trained multimodal reasoning capabilities of LVLMs to enable efficient, end-to-end multi-VIE without task-specific model training. As illustrated in Fig.~\ref{fig:LVLM_based}, the pipeline consists of three key components: (1) an image classification module to identify the document type, (2) an ICL-based prompt engineering module to construct task-specific prompts, and (3) an LVLM inference and post-processing module to generate structured predictions.

\subsubsection*{Classification}
The classification module serves two critical roles: rapidly filtering out non-target images and providing accurate document-type labels to guide downstream prompt construction, thereby reducing computational overhead and improving relevance.

We offer two practical implementation options:

\textbf{Option 1} (training-free) classifies images via feature similarity matching. Features are extracted from the input image using a pre-trained model, then matched against a reference feature index with cosine similarity, thresholding, and majority voting. The procedure is as follows:
\begin{itemize}    
 	\item \textbf{Feature Extraction:} $\mathbf{x}=M(I)$. A pre-trained model $M$ (such as ResNet) is utilized to extract the feature vector $\mathbf{x}$ from the input image $I$. 
    \item \textbf{Feature Vector Search:} $D, F = \mathrm{Index.search}(\mathbf{x}, K)$. A feature vector engine (e.g., Faiss) is used to identify the $K$ feature vectors most similar to $\mathbf{x}$. The resulting distance matrix $D \in \mathbb{R}^{1 \times K}$ and index matrix $F \in \mathbb{R}^{1 \times K}$ are then obtained.  
    \item \textbf{Similarity Calculation:} ${R_i} = \frac{\mathbf{x} \cdot \mathbf{y}_i}{|\mathbf{x}|\left|\mathbf{y}_i\right|}$, $\mathbf{y}_i=\mathrm{Index.reconstruct}(I_i)$. The cosine similarity between the input feature vector $\mathbf{x}$ and the reconstructed feature vector $\mathbf{y}_i$ is calculated for each search result $i$. 
    \item \textbf{Sorting:} $O = \mathrm{sort}({R})$. The similarity results are sorted in descending order, yielding the $K$ most similar image labels in the feature library, which are subsequently used to determine the final image category.  
    \item \textbf{Thresholding and Majority Voting}: A similarity threshold $\tau$ is applied to filter the sorted similarity scores $O$, retaining only those where $R_i \geq \tau$. From the retained results, the corresponding image labels are collected, and a majority voting scheme is employed: $C = \mathrm{mode}({L_j \mid R_j \geq \tau, j \in {1, \dots, K}})$, where $L_j$ is the label associated with the $j$-th reference vector, and $\mathrm{mode}$ selects the most frequent label.  
\end{itemize}

\textbf{Option 2} trains a supervised image classifier (e.g., EfficientNet \cite{tan2019EfficientNet} or ConvNeXt \cite{9879745_ConvNet}) on labeled samples, achieving higher accuracy at the cost of additional training.

Document-type identification is performed by a lightweight, training-free keyword-based classifier applied to OCR-extracted text:
\begin{equation}
\hat{d} = \arg \max_{d_i \in \mathcal{D}} \sum_{j=1}^{|\mathcal{T}|} \sum_{l=1}^{|\mathcal{K}(d_i)|} I(w_j, k_{il})
\end{equation}
where $\hat{d}$ is the predicted type, $\mathcal{T}$ is the set of recognized text tokens, $\mathcal{K}(d_i)$ is the predefined keyword set for type $d_i$, and $I(\cdot)$ is the indicator function. This simple yet effective mechanism matches distinctive terms (e.g., "risk assessment" vs. "emergency response" for similar certificate variants) and is easily extensible to new types by adding keyword sets.
 
\begin{figure}[H]
  \centering
  \includegraphics[width=1\linewidth]{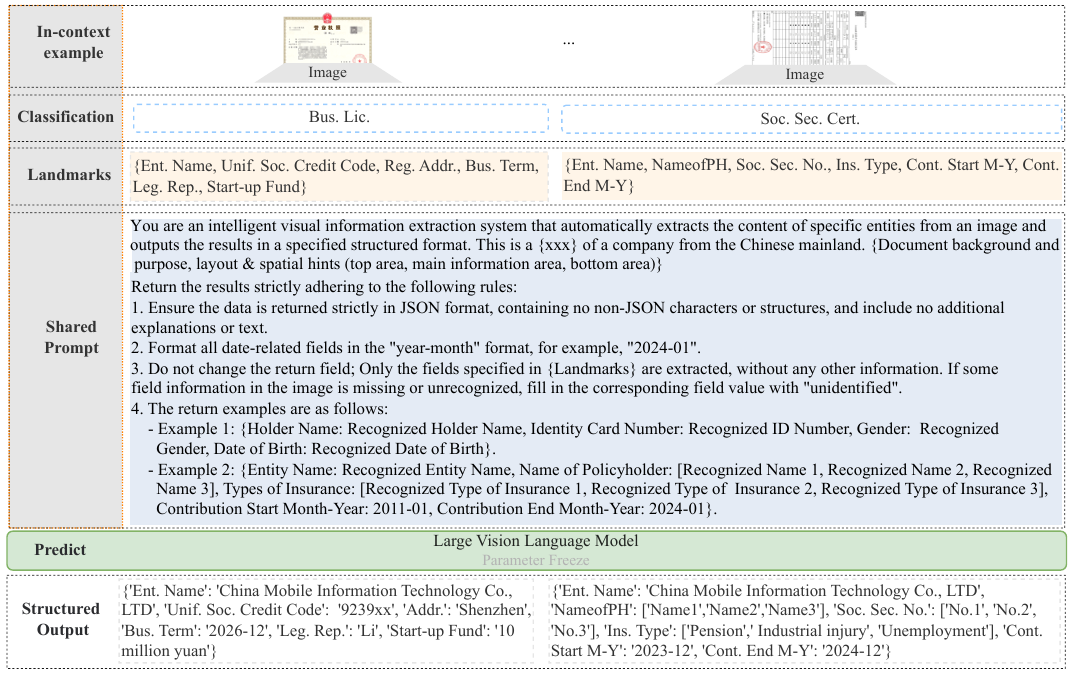} 
  \caption{Illustration of the dynamic prompt engineering process, showing shared instructions, type-specific landmarks, and in-context demonstrations.} 
  \label{fig:prompt} 
\end{figure}

\subsubsection*{ICL-Based Prompt Engineering} 
Once the document type is determined, we dynamically assemble a concise, task-specific prompt by combining a shared instruction block with type-specific components, as shown in Figure~\ref{fig:prompt}. The shared block defines the extraction task, provides general document background and purpose cues, describes common layout patterns and spatial hints (e.g., top/main/bottom areas), and enforces rigorous output rules: pure JSON format only, standardized date fields ("year-month"), extraction of specified landmarks exclusively, and "unidentified" for missing/unreadable values.

The predicted document type then injects the corresponding predefined entity list (landmarks) and appends 2--4 carefully selected in-context demonstrations. These demonstrations illustrate correct entity mapping, formatting adherence, and handling of missing information.

This classification-guided dynamic injection delivers highly relevant, focused context to the frozen LVLM, mitigating hallucinations caused by irrelevant or overloaded instructions while ensuring consistent, reliable extraction across diverse document layouts.

\subsubsection*{LVLM Inference and Post-Processing}
The constructed prompt and input image are fed to the LVLM, which generates structured predictions. Post-processing standardizes outputs (e.g., date formats, punctuation, capitalization) and ensures consistency for downstream applications such as database storage, knowledge graph construction, or intelligent retrieval.

This design enables robust adaptation to diverse document types, complex layouts, and real-world noise while maintaining high efficiency.

\subsubsection*{Domain-Specific Enhancement via Supervised Fine-Tuning}
For scenarios demanding maximum accuracy, we provide an optional enhancement through supervised fine-tuning of the base LVLM using low-rank adaptation (LoRA). Training incorporates multi-granular annotations: entity-level labels for fixed and variable fields, full-document OCR transcripts, and detailed image descriptions. This targeted adaptation significantly improves precision and further reduces hallucinations on the target domain while retaining much of the original model's generalization capability.

\subsection*{Theoretical Perspective on Classification-Guided Prompting}

In this section, we provide a theoretical perspective on why classification-guided prompt construction improves LVLM-based visual information extraction.

\paragraph{Problem Formulation.}
Let $I$ denote an input document image, $d \in \mathcal{D}$ its latent document type, and $Y$ the structured output. A LVLM performs conditional generation:
\begin{equation}
P(Y \mid I, P),
\end{equation}
where $P$ denotes the input prompt.

In conventional prompting, a universal prompt $P_{\text{all}}$ encodes instructions for all document types:
\begin{equation}
P_{\text{all}} = \bigcup_{d \in \mathcal{D}} P(d).
\end{equation}

In contrast, our method constructs a conditional prompt:
\begin{equation}
P = P(\hat{d}), \quad \hat{d} = g(I),
\end{equation}
where $g(\cdot)$ is the document classifier.

\paragraph{View as Conditional Computation.}
This formulation can be interpreted as a two-stage factorization:
\begin{equation}
P(Y \mid I) = \sum_{d \in \mathcal{D}} P(Y \mid I, P(d)) P(d \mid I).
\end{equation}

Our approach approximates this marginalization via a hard routing mechanism:
\begin{equation}
P(Y \mid I) \approx P(Y \mid I, P(\hat{d})),
\end{equation}
which is analogous to mixture-of-experts with a deterministic gating function.

This reduces the hypothesis space of the LVLM from all possible tasks to a task-specific subspace, improving sample efficiency in zero-shot settings.

\paragraph{Information-Theoretic Analysis.}
We analyze prompt effectiveness from an information perspective. Let $Z$ denote the token sequence of the prompt. The mutual information between prompt and output is:
\begin{equation}
I(Y; Z \mid I).
\end{equation}

A universal prompt $P_{\text{all}}$ contains both relevant and irrelevant information:
\begin{equation}
Z = Z_{\text{rel}} \cup Z_{\text{irr}}.
\end{equation}

Assuming irrelevant tokens are independent of the target output:
\begin{equation}
I(Y; Z_{\text{irr}} \mid I) \approx 0,
\end{equation}
but they still contribute to the entropy:
\begin{equation}
H(Z) = H(Z_{\text{rel}}) + H(Z_{\text{irr}}).
\end{equation}

Thus, the signal-to-noise ratio of the prompt can be defined as:
\begin{equation}
\text{SNR} = \frac{I(Y; Z_{\text{rel}} \mid I)}{H(Z)}.
\end{equation}

Classification-guided prompting removes $Z_{\text{irr}}$, yielding:
\begin{equation}
\text{SNR}_{\text{guided}} \gg \text{SNR}_{\text{all}},
\end{equation}
which leads to more efficient conditioning.

\paragraph{Attention Dilution Effect.}
Transformer-based LVLMs rely on attention mechanisms:
\begin{equation}
\text{Attn}(Q, K, V) = \text{softmax}\left(\frac{QK^T}{\sqrt{d}}\right)V.
\end{equation}

Let the prompt tokens be $\{z_1, \dots, z_n\}$. In long universal prompts, attention weights are distributed across both relevant and irrelevant tokens:
\begin{equation}
\sum_{i=1}^{n} \alpha_i = 1, \quad \alpha_i = \text{softmax}(q \cdot k_i).
\end{equation}

When $n$ increases due to irrelevant tokens, the expected attention mass assigned to relevant tokens decreases:
\begin{equation}
\mathbb{E}\left[\sum_{i \in \text{rel}} \alpha_i\right] \downarrow.
\end{equation}

We refer to this phenomenon as attention dilution, where useful signals are weakened by the presence of unrelated instructions.

\paragraph{Error Propagation Trade-off.}
The proposed method introduces a dependency on classification accuracy. Let $\epsilon = P(\hat{d} \neq d)$ denote classification error. Then:
\begin{equation}
P(Y \mid I) = (1 - \epsilon) P(Y \mid I, P(d)) + \epsilon P(Y \mid I, P(\hat{d} \neq d)).
\end{equation}

This reveals a trade-off:
\begin{itemize}
    \item High classification accuracy leads to strong gains via focused prompting,
    \item Misclassification introduces structured errors due to incorrect prompts.
\end{itemize}

Empirically, Table~\ref{tab:image_class} shows $\epsilon$ is small ($<2\%$), making the benefits dominant.

\paragraph{Connection to In-Context Learning.}
From an ICL perspective, prompt construction defines a task-specific context distribution:
\begin{equation}
P_{\text{context}} = P(Z \mid d).
\end{equation}

By conditioning on $d$, we align the context distribution with the input distribution:
\begin{equation}
P(Z \mid d) \approx P(Z \mid I),
\end{equation}
which reduces distribution mismatch and improves generalization.

Overall, classification-guided prompting improves LVLM performance by reducing the hypothesis space through conditional routing, increasing the signal-to-noise ratio of prompts, mitigating attention dilution in long prompts, and better aligning the context distribution for in-context learning. These factors together provide a theoretical foundation for the empirical gains observed in our experiments.

\subsection*{Alternative OCR\&UIE Pipeline as Baseline}
\begin{figure*}   
  \centering
  \includegraphics[width=1\linewidth]{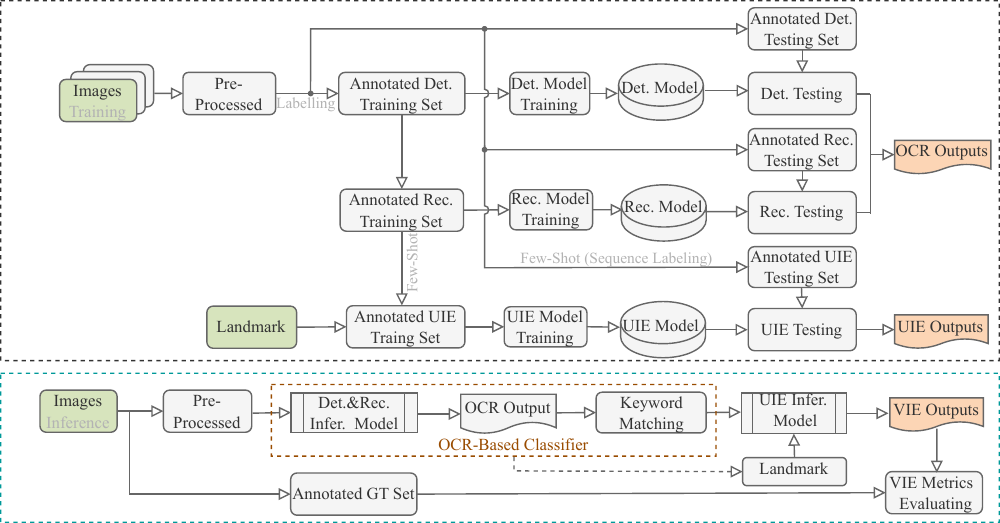} 
  \caption{Overview of the OCR\&UIE baseline pipeline, incorporating a keyword-based classifier, pre-trained OCR, and per-type UIE models.
  }
  \label{fig:ocr-uie} 
\end{figure*}

To establish a strong trainable baseline, we develop an enhanced two-stage pipeline that integrates mature OCR with unified information extraction (UIE), as depicted in Figure~\ref{fig:ocr-uie}.

The UIE module replaces traditional semantic entity recognition, requiring only a small number of annotated samples (e.g., 10 per type) for robust performance. Combined with mature pre-trained OCR models, this pipeline offers a practical solution for small-sample multi-VIE scenarios.

\section*{Experiment} 
This study evaluates the effectiveness of the proposed method using a real‑world electronic bidding and tendering platform. The system handles extensive document‑based information exchange, where a key challenge lies in accurately extracting and analyzing information from image‑based documents. These documents include, but are not limited to, business licenses, professional qualification certificates, and social security certificates. They serve two critical purposes: (1) as legal proof of compliance and eligibility for bidders, and (2) as fundamental data for tendering entities to perform qualification reviews and risk assessments.

\subsection*{Dataset} 
Public benchmarks such as SROIE~\cite{huang2019icdar2019} and SCID~\cite{SCID} are limited in document variety, layout complexity, and entity types, and may have been exposed during large vision-language model pretraining. To enable rigorous evaluation under realistic conditions, we construct a new domain-specific dataset collected from a real-world electronic bidding and tendering platform.

The dataset comprises 98,600 images of 16 common certificate types, sourced from production environments across diverse regions, industries, and enterprise scales. These samples exhibit authentic real-world variations, including differences in resolution, watermarks, seal imprints, and photographic distortions. Original documents were provided in PDF, PPT, XLS, DOC, and image formats; embedded images were extracted using PyMuPDF, pdf2image, and Pillow, followed by preprocessing with OpenCV.
The data were split into training and test sets in a 7:3 ratio. Due to commercial sensitivity, the original images are not publicly released; however, representative synthetic examples generated via GPT-4V are provided in our open-source repository for illustrative purposes.

Table~\ref{tab:dataset_stats} summarizes the per-category distribution and key characteristics. The dataset maintains reasonable balance across categories while exhibiting substantial variation in extraction complexity: the average number of target entities ranges from 4 to 11 (overall 8.5), and OCR-extracted text length varies significantly (overall 312.81 characters), with particularly dense tabular content in categories like Social Security Certificates. 
\begin{table}[t]
\centering
\caption{Statistics of the real-world bidding dataset.}
\label{tab:dataset_stats}
\begin{tabular}{|l|c|c|c|c|c|c|}
\hline
No. & Document Type & Train Samples & Test Samples & Avg. Entities  & Avg. Text Length \\
\hline
\hline
01 & Acad. Qual. Cert.        & 5156  & 2210  & 11 & 262 \\ \hline
02 & Deg. Cert.               & 4456  & 1910  & 8 & 227 \\ \hline
03 & PILPC                    & 4209  & 1804  & 8 & 254 \\ \hline 
04 & Bus. Lic.                & 6112  & 2619  & 8 & 453 \\ \hline 
05 & Soc. Sec. Cert.          & 6233  & 2671  & 8 & 893 \\ \hline 
06 & ID Card                  & 2927  & 1254  & 4 & 125 \\ \hline 
07 & ISO 9001 QMS Cert.       & 4156  & 1781  & 8 & 250 \\ \hline 
08 & ISO 14001 EMS Cert.      & 3691  & 1582  & 8 & 246 \\ \hline 
09 & SA 8000 Cert.            & 4328  & 1855  & 8 & 244 \\ \hline
10 & ISO 45001 OHSMS Cert.    & 4056  & 1738  & 8 & 231 \\ \hline 
11 & CSCRC                    & 4390 &  1881  & 8 & 307 \\ \hline 
12 & TNSSCC (Risk Assess.)    & 3789  & 1624  & 9 & 290 \\ \hline 
13 & TNSSCC (Emerg. Resp.)    & 4012  & 1719  & 9 & 319 \\ \hline 
14 & TNSSCC (Des. \& Integr.) & 3812  & 1634  & 9 & 328 \\ \hline 
15 & TNSSCC (Sec. Train.)     & 4141  & 1775  & 9 & 336 \\  \hline
16 & PCI                      & 3552  & 1523  & 9 & 240 \\ \hline    
\multicolumn {2}{|c|} {Total} & 69,020 & 29,580 & 8.5  & 312.81 \\  \hline
\end{tabular}
\end{table}

\subsection*{Model Training} 
For the detection and recognition tasks, we selected the differentiable binarization algorithm, which facilitates efficient post-processing, and the convolutional recurrent neural network algorithm, which integrates convolutional and sequential features. Notably, accuracy was enhanced by fine-tuning a pre-trained model from PaddleOCR \cite{li2022pp}, which demonstrated optimal accuracy and generalization performance for publicly available Chinese-language datasets. For the UIE task, following the approach outlined in \cite{lu2022unified_UIE}, separate UIE models were trained for each image type to extract the corresponding landmark fields. 
In the OCR-based classification strategy, a pre-trained ResNet-101 model extracts deep features from pre-processed images, with the final fully connected layer removed to produce feature vectors.  Image preprocessing involves resizing to \( 400 \times 300 \) pixels, converting to tensor format, and normalizing using mean \([0.485, 0.456, 0.406] \) and standard deviation \( [0.229, 0.224, 0.225] \). For query image classification, cosine similarity is computed against a Faiss-based index, and top-5 nearest neighbors are retrieved with a thresholding mechanism (\( \tau = 0.9 \)) and majority voting to enhance prediction reliability. For the detection and recognition labeling tasks, the PaddleLabel tool \cite{paddlelabel2022} was employed, while the Doccano tool \cite{doccano} was utilized for sequence labeling in the UIE task. 
The detection, recognition, and UIE models were trained using a single NVIDIA 3090 GPU. For image classification, we utilized the ConvNeXt-B ImageNet-22K 224 configuration \cite{woo2023convnext} pre-trained model, which was trained on a single NVIDIA 3090 GPU.

To further reduce the inference time of LVLM while improving accuracy, we fine-tuned the Qwen2.5-VL-7B model using LoRA training.  We first performed preliminary annotation on the raw data using a large-scale LVLM (Qwen2.5-VL-72B), followed by manual verification and correction of errors. 
A total of 394,400 pieces of annotated training data were compiled for the experiment. 
These four types of annotated data not only enhance the model’s understanding of document information from different perspectives but also help mitigate the issue of reduced generalization ability after fine-tuning. The training was conducted using 8 NVIDIA A100 GPUs over a period of 6 days.

\subsection*{Evaluation Metrics}  
In the context of image classification, let $\mathcal{L}$ represent the set of all possible labels. For each label \(l \in \mathcal{L}\), we define \(N_t^{l}\) as the number of true positives, \(N_f^{l}\) as the number of false positives, and \(N_n^{l}\) as the number of false negatives. The F1-score for each $l$ is computed as follows: 
\begin{equation}
\label{eq:f1}
\mathrm{F1}^{l} = 2 \times \frac{\mathrm{Precision}^{l} \times \mathrm{Recall}^{l}}{\mathrm{Precision}^{l} + \mathrm{Recall}^{l}}
\end{equation}
where $\mathrm{Precision}^{l} = \frac{N_t^{l}}{N_t^{l}+N_f^{l}}$ and $\mathrm{Recall}^{l} = \frac{N_t^{l}}{N_t^{l}+N_n^{l}}$. 
The weight \(w^{l}\) assigned to each label \(l\) is given by:
\begin{equation}
w^{l} = \frac{N_t^{l}+ N_n^{l}}{\sum_{l \in \mathcal{L}} (N_t^{l}+ N_n^{l})}
\end{equation}
The weighted-average F1-score is then computed as:
\begin{equation}
\mathrm{F1_{weighted}} = \sum_{l\in\mathcal{L}}w^{l} \mathrm{F1}^{l} 
\end{equation}

For the VIE task, predictions are compared with ground truth labels, and both entity-level F1-scores and the average normalized edit distance (NED) are used to evaluate performance \cite{VIE_wang2021towards, GPT_4V_shi2023exploring}. An entity is considered a true positive if its predicted content exactly matches the corresponding ground truth.
The $\mathrm{Precision}$ and $\mathrm{Recall}$ are defined as:  
\begin{equation}
\label{eq:vie}
\left\{
\begin{aligned} 
& \mathrm{Precision} = \frac{N_t}{N_p} \\
& \mathrm{Recall} = \frac{N_t}{N_g}
\end{aligned}
\right.
\end{equation}
where \(N_t\) denotes the number of true positive samples, \(N_p\) represents the total number of predictions, and \(N_g\) is the number of ground truth instances. The entity-level F1-score is computed by combining equations \eqref{eq:vie} and \eqref{eq:f1}.

Moreover, the average NED-score is given by:
\begin{equation}
\begin{aligned}
\mathrm{NED} = \frac{1}{N} \sum_{k=1}^N  \left\{ 1 - \frac{I_k + D_k + M_k}{L_{g,k}} \right\}  
\end{aligned}
\end{equation}   
where \(N\) represents the total number of entities, and \(I_k\), \(D_k\), \(M_k\), and \(L_{g,k}\) denote the number of insertions, deletions, modifications of the \(k\)-th entity, and the total number of instances occurring in the ground truth, respectively.

\subsection*{Results and Discussion}  
\subsubsection*{Classification Performance}
To evaluate the proposed classification framework, we conducted experiments under four configurations:
\begin{itemize}
    \item Retrieved Feature Matching (RFM) (classification Option 1): A baseline method using a deep residual network to extract image features, followed by cosine-similarity matching against a database. It serves as a visual-only reference.
    \item Feature Matching-Text Fusion (FM-TF) (classification Option 1 with OCR-based classifier): Extends RFM by integrating an OCR module to extract textual information, thereby refining classification using both visual and textual cues.
    \item Trained End-to-End Multi-Classification (TEEMC) (classification Option 2 with multi-classification mode): Employs a lightweight ConvNeXt-B network pre-trained on ImageNet-22K and fine-tuned on our dataset in multi-class mode, enhanced with an OCR-based module.
    \item Trained End-to-End Binary-Classification (TEEBC) (classification Option 2 with binary-classification mode): Similar to TEEMC but uses multiple independent binary classifiers, each dedicated to a specific class.
\end{itemize}

\begin{table}[H] 
  \centering
  \caption{F1-scores for the classification tasks.}   
  \label{tab:image_class} 
  \begin{tabular}{|c|c|c|c|c|}
    \hline
     Method & RFM & FM-TF & TEEMC & TEEBC  \\
    \hline\hline
     $\mathrm{F1_{weighted}}$ & 78.01\% & 91.41\% & 82.37\% & 98.45\% \\  
    \hline
  \end{tabular} 
\end{table}

As shown in Table~\ref{tab:image_class}, the RFM method, relying solely on visual features, yields the lowest F1-score. In contrast, FM-TF, which incorporates OCR-extracted text, improves accuracy substantially. This fusion addresses a key challenge in supply chain bid evaluation: distinguishing visually similar certificates (e.g., Telecommunication Network Security Service Capability Certificates) that differ only in sparse keywords (e.g., “risk assessment,” “emergency response”). The end-to-end trained models TEEMC and TEEBC achieve further gains, with TEEBC attaining the highest performance (98.45\%). This indicates that dedicated binary classifiers outperform a single multi-class model on our dataset.
 
In summary, combining end-to-end training with textual feature fusion is crucial for high classification accuracy. While the feature-matching approach (Option~1) offers good scalability without extra labeled data, the training-based approach (Option~2) delivers superior effectiveness. Method selection in practice should therefore align with task-specific demands and constraints.

\subsubsection*{Multi-VIE Performance}
We compare our fine-tuned 7B model against two baseline versions of Qwen2.5-VL (7B and 72B) on a general document information extraction task. Results are shown in Table~\ref{tab:multi-vie}.  

\begin{table}[h]
  \centering
  \caption{F1-score and NED for multi-VIE.} 
  \label{tab:multi-vie} 
  \begin{tabular}{|c|c|c|c|c|}
    \hline
     Model & Qwen2.5-VL-72B & Qwen2.5-VL-7B & Ours  \\
    \hline\hline
     $\mathrm{F1}$ & 86.98\% & 86.43\% & 93.65\% \\  
     \hline
     $\mathrm{NED}$ & 0.9037  & 0.9012 & 0.9348 \\   
    \hline
  \end{tabular}
\end{table}

Our fine-tuned 7B model outperforms both the untuned 7B and the larger 72B model. 
In addition, we also measured inference latency to assess the practical deployability. The Qwen2.5-VL-72B model requires an average of 6.5 seconds per image, while both the base Qwen2.5-VL-7B and our fine-tuned 7B model average 2.6 seconds per image. This approximately 2.5× speedup of the 7B models, combined with their superior or comparable accuracy on our real-world dataset, demonstrates a favorable trade-off for real-world deployment scenarios where throughput and resource constraints are critical. 

\subsubsection*{Cross-Dataset Generalization Analysis}

To evaluate generalizability, we conduct experiments on multiple public benchmarks with diverse layouts and domains (Table~\ref{tab:bench}). 

The results show that our method maintains competitive performance across datasets without task-specific adaptation. While slight performance degradation is observed on datasets such as FUNSD and EPHOIE, this is expected due to domain shift in layout structure and annotation schema.

Notably, the performance gap remains relatively small compared to base LVLMs, indicating that classification-guided prompting does not overfit to the source domain. Instead, it preserves the general reasoning capability of LVLMs while improving task alignment.
These results demonstrate that the proposed framework generalizes well across datasets and is robust to distribution shifts, which is critical for real-world deployment.

\begin{table}[h]
  \centering
  \caption{Benchmark results (F1 / NED) on public datasets.} 
  \label{tab:bench} 
  \begin{tabular}{|c|c|c|c|c|}
    \hline
     Dataset & \makecell{Qwen2.5-VL-72B \\ F1 / NED } & \makecell{Qwen2.5-VL-7B \\ F1 / NED}  & \makecell{Ours \\ F1 / NED}  \\
    \hline\hline
    COLD-CELL & 79.73\% / 0.9451 & 79.8\% / 0.9439 & 78.13\% / 0.9370 \\
    COLD-SIBR & 76.22\% / 0.8936 & 76.2\% / 0.8929  & 75.12\% / 0.8839  \\ 
    EPHOIE & 85.22\% / 0.9365 & 85.24\% / 0.9358 & 81.55\% / 0.9214  \\
    FUNSD & 27.64\% / 0.4186 & 27.44\% / 0.4173 & 26.88\% / 0.4029  \\
    ICDAR & 83.8\% / 0.9444 & 83.5\% / 0.9416  & 84.49\% / 0.9373  \\ 
    POIE & 86.99\% / 0.9439 & 87.16\% / 0.9415  & 86.15\% / 0.9359  \\
    SROIE & 83.74\% / 0.9401 & 83.86\% / 0.9393  & 83.18\% / 0.9299  \\ 
    XFUND-zh & 59.13\% / 0.6337 & 58.35\% / 0.6420 & 60.35\% / 0.6578  \\ 
    \hline
  \end{tabular}
\end{table}

\subsubsection*{OCR\&UIE vs. Classification-Guided LVLM}
Table~\ref{tab:performance} compares the conventional OCR\&UIE pipeline with our proposed classification-guided LVLM approach across 16 certificate types. 
\setlength{\tabcolsep}{2.5mm}
\begin{table*}[h]
  \centering
  \caption{F1 and NED comparison between OCR\&UIE and LVLM-based multi-VIE.}
  \label{tab:performance}
  \begin{tabular}{|>{\centering\arraybackslash}m{1cm}|>{\centering\arraybackslash}m{3.8cm}|>{\centering\arraybackslash}m{1.8cm}|>{\centering\arraybackslash}m{1.6cm}|>{\centering\arraybackslash}m{1.8cm}|>{\centering\arraybackslash}m{1.6cm}|}
    \hline
    No. & Document Type & OCR\&UIE-based F1  & LVLM-based F1 & OCR\&UIE-based NED & LVLM-based NED \\
    \hline\hline
    01 & Acad. Qual. Cert. & 70.58\% & 95.90\% &0.6512 &0.8770 \\
    \hline
    02 & Deg. Cert. & 65.86\% & 92.15\% &0.6303 &0.8617 \\
    \hline
    03 & PILPC & 51.11\% & 85.23\% &0.6127 &0.9022 \\
    \hline
    04 & Bus. Lic. & 70.85\% & 94.44\% &0.8021 &0.9903 \\
    \hline
    05 & Soc. Sec. Cert. & 43.65\% & 88.02\% &0.3200 &0.7635 \\
    \hline
    06 & ID Card & 91.67\% & 99.36\% &0.8917 &0.9821 \\
    \hline
    07 & ISO 9001 QMS Cert. & 66.67\% & 97.59\% &0.7456 &0.9952 \\
    \hline
    08 & ISO 14001 EMS Cert. & 62.32\% & 96.30\% &0.6546 &0.9709 \\
    \hline
    09 & SA 8000 Cert. & 70.20\% & 92.67\% &0.7092 &0.9617 \\
    \hline
    10 & ISO 45001 OHSMS Cert. & 61.11\% & 97.12\% &0.6873 &0.9680 \\
    \hline
    11 & CSCRC & 69.94\% & 97.42\% &0.7008 &0.9754 \\
    \hline
    12 & TNSSCC (Risk Assess.) & 77.92\% & 93.47\% &0.6474 &0.9649 \\
    \hline
    13 & TNSSCC (Emerg. Resp.) & 69.18\% & 91.84\% &0.6243 &0.9480 \\
    \hline
    14 & TNSSCC (Des. \& Integr.) & 75.57\% & 90.89\% &0.7142 &0.9221 \\
    \hline
    15 & TNSSCC (Sec. Train.) & 72.38\% & 91.43\% &0.7123 &0.9553 \\
    \hline
    16 & PCI & 70.34\% & 94.62\% &0.7034 &0.9609 \\
    \hline 
    \multicolumn {2}{|c|}{Average} & 68.08\% & {93.65\%} & 0.67 & {0.9348} \\
\hline
  \end{tabular}
\end{table*}
  
The classification-guided LVLM-based method consistently outperforms OCR\&UIE, with average F1 improving from 68.08\% to 93.65\% and average NED from 0.67 to 0.93. The gain is especially pronounced for form-like documents (e.g., Social Security Certificate, +44.37\% F1). By combining the results in Table~\ref{tab:multi-vie} and Table~\ref{tab:performance}, it can be observed that even without any fine-tuning, the LVLM-based method has improved by 18.35\% in the average F1-score and by 0.23 in the average NED. 
Despite the fine-grained fine-tuning employed in the OCR\&UIE-based method, its performance remains suboptimal in real-world multi-VIE scenarios. This limitation stems from the inherent dependency of VIE accuracy on the outputs of detection and recognition stages. In practical applications, results are influenced not only by text attributes, such as color, size, font, shape, orientation, and multilingual content, but also by various image distortions, including blurring, low resolution, shadows, brightness variations, watermarks, and seal obstructions. 
In contrast, the LVLM-based approach benefits from extensive pre-training on large-scale image-text datasets, providing a robust understanding of image elements and enhanced predictive capabilities for low-quality text information. 

\subsubsection*{Ablation Study}
To validate the effectiveness of key components, we conduct ablation experiments on the test set using the zero-shot Qwen2.5-VL-7B model.

\begin{table}[h]
\centering
\caption{Ablation study on core components (zero-shot setting).}
\label{tab:ablation}
\begin{tabular}{|l|c|c|}
\hline
Method & F1-score (\%) & NED \\
\hline\hline
Full Framework (Classification + ICL)       & 86.43 & 0.9012 \\ 
- w/o Classification (single universal prompt) & 68.42 & 0.7312 \\ 
- w/o ICL (only task definition + format)    & 79.65 & 0.8431 \\ 
- w/o Post-processing                        & 84.97 & 0.8876 \\ 
- w/o Image Rotation Preprocessing           & 83.21 & 0.8723 \\ \hline 
Classification Option 1 (feature matching)   & 85.79 & 0.8956 \\ 
Classification Option 2 (trained ConvNeXt)   & 86.43 & 0.9012 \\ \hline 
\end{tabular}
\end{table}

As shown in Table~\ref{tab:ablation}, removing document-type classification causes the most significant performance drop (18.01 percentage points in F1-score). In this ablated setting (``single universal prompt''), we construct one monolithic prompt that includes landmarks, background descriptions, layout hints, and in-context examples for \emph{all} 16 document types simultaneously (see Figure~\ref{fig:prompt} for the per-type structure). This results in an extremely long prompt, forcing the LVLM to infer the correct document type and corresponding extraction rules entirely from the input image and the overloaded prompt context. The substantial degradation is likely due to diluted attention: the model struggles to focus on the relevant subset of instructions amid the large volume of irrelevant information for the given document type.

The removal of ICL (``only task definition + format'') also notably harms performance (+6.78 percentage points F1 contribution from ICL). Here, we retain document-type classification and dynamically select the correct landmarks but omit all in-context examples and task-specific knowledge injection (e.g., document background and purpose, layout \& spatial hints such as ``top area, main information area, bottom area'' illustrated in the shared prompt component of Figure~\ref{fig:prompt}). Without these structured demonstrations and prior knowledge, the zero-shot LVLM lacks sufficient guidance to reliably locate and extract entities in visually rich, domain-specific certificates, leading to increased hallucinations and formatting errors.

Post-processing and image rotation preprocessing provide smaller but consistent gains, confirming their value in handling real-world variations (e.g., inconsistent formatting and vertically oriented scans). Finally, the trained classifier (Option 2) slightly outperforms the training-free feature-matching approach (Option 1), indicating that higher classification accuracy directly benefits downstream extraction in the zero-shot pipeline.

\subsubsection*{Limitations and Mitigations}

Although the proposed classification-guided LVLM framework demonstrates strong performance and robustness in real-world multi-VIE scenarios, it inherits several limitations common to generative vision-language models.

The primary issue is hallucination, where the model generates plausible but incorrect or incomplete entity values. This problem is particularly pronounced in documents with degraded visual quality (e.g., blur, low contrast, or occlusion), where the model tends to rely on prior knowledge rather than faithfully grounding its predictions in the input.

We observe several characteristic failure modes:

\begin{itemize}
    \item Long-text instability: When extracting fields with long textual content (e.g., addresses or descriptions), the LVLM may produce degenerate outputs, including repetitive token generation. In extreme cases, the model repeatedly outputs a single word or phrase until reaching the token limit, indicating instability in long-sequence decoding.
    
    \item Visual ambiguity: Documents with highly similar layouts but subtle semantic differences (e.g., certificate variants) may lead to incorrect field assignments, especially when classification confidence is low.
    
    \item Error propagation from classification: Since prompt construction depends on predicted document type, misclassification can result in systematically incorrect prompts and structured extraction errors.
\end{itemize}

To mitigate these issues, we adopt several practical strategies. First, automatic rotation preprocessing is applied to align images horizontally, yielding an approximately 3 percentage point improvement in average F1-score. Second, post-processing enforces format normalization and removes obvious repetition artifacts in generated outputs. 

For more fundamental improvements, future work could explore constrained decoding strategies (e.g., JSON-schema-guided generation, repetition penalties, or length-aware stopping criteria) to stabilize long-text generation. Additionally, improving classification confidence estimation and incorporating soft or multi-hypothesis prompting may further reduce error propagation.

\section*{Conclusion}   
We propose a classification-guided LVLM framework for real-world multi-VIE tasks in visually rich documents. By decoupling document-type identification from content extraction and introducing dynamic prompt-based knowledge injection, our approach achieves superior efficiency and generalization without task-specific training in its base form. 
On a challenging real-world bidding dataset comprising 16 diverse certificate types, the zero-shot LVLM configuration substantially outperforms the trainable OCR\&UIE baseline, improving average F1-score from 68.08\% to 86.43\% and normalized edit distance from 0.67 to 0.90. Optional domain-specific fine-tuning further elevates performance to 93.65\% F1-score and 0.93 normalized edit distance, demonstrating remarkable robustness against real-world impairments such as seals, watermarks, and low contrast.
This work establishes an accessible, scalable paradigm for complex document understanding, offering a practical evolutionary path for intelligent document processing systems in office automation and beyond. 
More broadly, our findings suggest that input-dependent prompt conditioning can be viewed as an effective approximation to conditional computation in large vision-language models, highlighting a promising direction for improving information efficiency and controllability in prompt-based multimodal inference.

\section*{Acknowledgements}
This work was supported by the Ningbo Natural Science Foundation under Grant 2023J285.

\section*{Author contributions statement}
All authors contributed to the study conception and design. Material preparation, data collection, and analysis were conducted by Huafu Li, Guo Chen, Jia Xia, and Liming Li. Lei Wang was responsible for statistical analysis and validation of data interpretation. Wei Du oversaw experimental methodology development and quality control during data collection. Yun Yao and Weijun Peng participated in manuscript revision, provided critical review of the scientific content, and approved the final version for publication. The first draft of the manuscript was written by Huafu Li, and all authors reviewed and commented on subsequent versions. All authors read and approved the final manuscript.

\section*{Additional information}
\begin{itemize}
\item Competing interests: The authors have no competing interests to declare.
\item Ethics approval and consent to participate: Not applicable.
\item Consent for publication: Not applicable.
\item Funding: This work was supported by the Ningbo Natural Science Foundation under Grant 2023J285.
\item Data availability: The source code is publicly available
on GitHub at https://github.com/FairmeHIT/Multi-VIE, and the fine-tuned models can be accessed via Hugging Face at
https://huggingface.co/fairme/Qwen2.5-VL-7B-SFT.
\end{itemize}

The corresponding author is responsible for submitting a \href{http://www.nature.com/srep/policies/index.html#competing}{competing interests statement} on behalf of all authors of the paper. This statement must be included in the submitted article file.

\bibliography{sn-bibliography} 

\end{document}